\documentclass{article}
\usepackage{spconf,amsmath,graphicx}
\usepackage{multirow}
\usepackage{algorithm}
\usepackage{array,booktabs}
\usepackage{url}
\usepackage{algorithm}
\usepackage[noend]{algpseudocode}
\usepackage{todonotes}
\usepackage{amssymb}

\title{Efficient Semi-Supervised Learning for Natural Language Understanding by Optimizing Diversity}
\name{Eunah Cho$^1$, He Xie$^1$, John P. Lalor$^{2*}\thanks{* Work done during an internship at Amazon Alexa}$, Varun Kumar$^1$, William M. Campbell$^1$}
\address{$^1$Amazon Alexa AI Natural Language Understanding\\
  $^2$University of Notre Dame}

\begin{document}
%
\maketitle
\begin{abstract}
Expanding new functionalities efficiently is an ongoing challenge for single-turn task-oriented dialogue systems. In this work, we explore {\it functionality-specific} semi-supervised learning via self-training. We consider methods that augment training data automatically from unlabeled data sets in a functionality-targeted manner. In addition, we examine multiple techniques for efficient selection of augmented utterances to reduce training time and increase diversity. First, we consider paraphrase detection methods that attempt to find utterance variants of labeled training data with good coverage. Second, we explore sub-modular optimization based on $n$-grams features for utterance selection. Experiments show that functionality-specific self-training is very effective for improving system performance. In addition, methods optimizing diversity can reduce training data in many cases to $50\%$ with little impact on performance.

\end{abstract}

\section{Introduction}
Single-turn task-oriented dialogue systems have become commonplace. These systems are based on spoken language understanding which takes speech input 
and processes it 
with natural language understanding (NLU) to produce intents, domains, and slots~\cite{tur2011spoken}. As an example, a dialog system may have the input ``set an alarm for five p.m. tomorrow'' and produce domain \textit{alarm}, intent \textit{set alarm}, and slot values of ``five p.m.'' for \textit{time} and ``tomorrow'' for \textit{date}. In this work we define a functionality as one or two NLU intents with their associated slots, that our dialog system can group into one model capability (e.g. ``SettingAlarm"). 

One of the ongoing challenges with production applications is to quickly design and add new functionalities. Building accurate NLU models for new functionalities requires collection and manual annotation of new data which is an expensive process. Semi-supervised learning (SSL), learning from both unlabeled and existing labeled data, potentially provides a low-cost yet efficient method to improve NLU models performance. 

Maintaining training data so that it is relevant with current usage pattern as well as to achieve efficient training is another challenge in production applications. As new functionalities are being introduced, it is expected to have constant shift in usage pattern. At the same time, continuously adding data for each functionality increases complexity for training.  

For this work, we focus on self-training for semi-supervised NLU that provides performance improvement for a specific functionality. In self-training, the system improves its performance by applying the current NLU model to unlabeled utterances, selecting functionality-relevant data, and then augmenting this data back into the system training set. As new system capabilities emerge, self-training can be focused on the functionality of interest and improve system performance quickly with no human annotation effort. In contrast, other semi-supervised methods may provide overall system performance improvements, but yield no performance improvements for a particular functionality.
In the following discussion, we explore a method with diversity measures to sub-sample data in order to achieve efficient training with SSL and make the model agile to shifts in usage pattern.  

Popular SSL models include self-training, mixture models, graph-based methods, co-training and multiview learning ~\cite{zhu2005semi, zhu2009introduction, chapelle2010ssl}. Particularly, in speech and language processing, self-training~\cite{ma2006unsupervised,tur2005combining,mcclosky2006effective,reichart2007self} and methods that learn representations from implicit information are common~\cite{collobert2008unified,peters2018deep}. In \cite{oliver2018realistic}, authors shared their insights on how evaluation of SSL approaches should be made when using production settings. 
Diversity in SSL was considered for text categorization tasks \cite{li2013semi}. In this work, diversity is used to better exploit unlabeled data by using feature subspace classifiers. In our work, we use diversity as a criterion to decrease the amount of training data necessary for maintaining the NLU performance. 
The authors in \cite{wieting2015towards} investigated six models for paraphrase embedding, varying in terms of expressiveness and complexity including deep averaging network \cite{iyyer2015deep} and LSTM networks \cite{hochreiter1997long}. Inspired by this work, we examined the six models for embedding and selected the best performing one for our experiments.

Prior approaches to self-training for NLU \cite{wu2010spoken, gotab2010online} have primarily focused on model performance. In this paper, we focus on optimizing both NLU model performance and augmented training data diversity. As the unlabeled data pool grows and functionality usage increases, the amount of functionality-specific data can grow arbitrarily. Much of the variation in this data is minimal and does not improve system performance. Thus, the challenge with functionality-specific SSL is to find the functionality in the unlabeled pool, minimizing false positives given limited training data. 

In addition, incorporation of large amounts of training data can linearly increase training times with neural network and embedding-based techniques. To address these concerns, we optimize augmented training set size with a selection criterion.
Efficiently learning new functionalities with less data is a common problem across many modern voice assistants. Methods for optimizing measures similar to diversity are seen in multiple areas including general deep learning, e.g. active learning in NLU \cite{sener2017active, shen2017deep}. In order to ensure external validity of our experiments, we selected NLU functionalities with varying number of slots and intent combination.

Our contribution in this paper is two-fold. First, we show that {\it functionality-specific} semi-supervised learning substantially improves production system performance for new capabilities, reaching up to 25\% in relative performance difference. Second, we explore different methods for optimizing the size of the augmented training data set. Using a paraphrase detection model, we propose a greedy algorithm for optimizing diversity in the augmented data set. In addition, we compare our proposed paraphrase model with sub-modular data selection method. We demonstrate that applying these techniques to the unlabeled training data yields substantially smaller training data sets (down to 50\%) with comparable performance to unrestricted augmentation. 


\section{Semi-Supervised Learning for NLU}\label{ssl-nlu}

\begin{figure}[th]
    \centering
    \includegraphics[width=7cm]{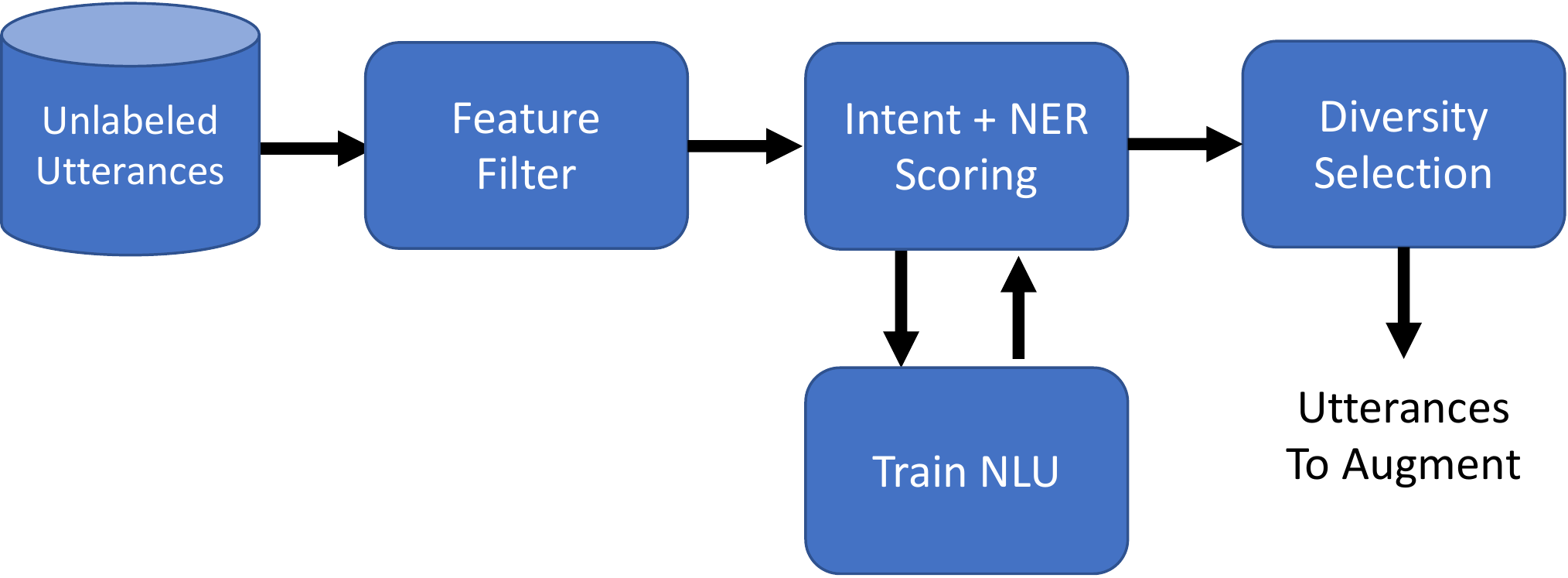}
    \caption{SSL using self-training and diversity}
    \label{fig:ssl}
\end{figure}

Our approach to SSL is shown in Figure~\ref{fig:ssl}. For the first step, we take a pool of unlabeled utterances and apply a functionality filter. In order to simulate the SSL process, we gathered live traffic utterances from 2017 January to 2018 June, and used these utterances as the pool for SSL utterances. 

The functionality filter is a 1-vs-rest classifier trained with in-class examples of the functionality and all other functionalities as out-of-class. A low complexity $n$-gram based linear logistic regression classifier is used for our implementation for high-throughput and to reduce volume. As features, we used unigram, bigram and trigram of the token sequences. Only utterances above a threshold are examined for subsequent stages. In this experiment, we used $0.5$ as a fixed threshold.

%

For the next step in the figure, we iteratively refine our selection by applying both intent classification (IC) and named entity recognition (NER) to the filtered utterances. A fused system score is obtained by multiplying confidence score from each NLU component (e.g. IC) and 
used for selection of candidate utterances above a threshold. Given that the confidence score is normalized between $0$ and $1$, we experimented with a varying range of confidence score thresholds, $\{0.2, 0.3, 0.4, ..., 0.9\}$. The selected utterances are augmented back into training, where the data is weighted same as the threshold score. Thus, the more the model is confident, the more important the data is for training. In our preliminary experiments, we learned that we can achieve the best performance in SSL when we aggregate all data over different thresholds and add the aggregated data with a constant weight $1.0$. Since highly confident utterances are included in the data set with a lower threshold, this data is already weighted towards more confident data. We will be using this aggregating method for SSL in this work.  

As a final step, we take the results of the first two stages, namely the aggregated data, and apply a diversity selection process. Since the initial stages are confidence based, the output size can be arbitrary based on the size of the unlabeled data pool, the popularity and newness of the functionality, and the precision/recall of the filter. The diversity process ensures we obtain a parsimonious final set of utterances to augment into the final training process. Detailed description on the diversity based selection will be given in Section \ref{paraphrase-detection}. 

\section{Diversity in SSL}
Our intuition of the SSL process is that we can train the NLU model more efficiently by prioritizing utterances with diversity. Utterances with diversity will introduce lexical and syntactic variety into the training data. For example, let us assume that the model has already seen an utterance \textit{play Adele} and thus can support the feature for the given carrier phrase. Compared to prioritizing the same or very similar utterances, prioritizing utterances with more variety (e.g. \textit{I want to listen to Lady Gaga}) into training would help enhance the model to support the target feature.   
\begin{figure}[th]
    \centering
    \includegraphics[width=6cm]{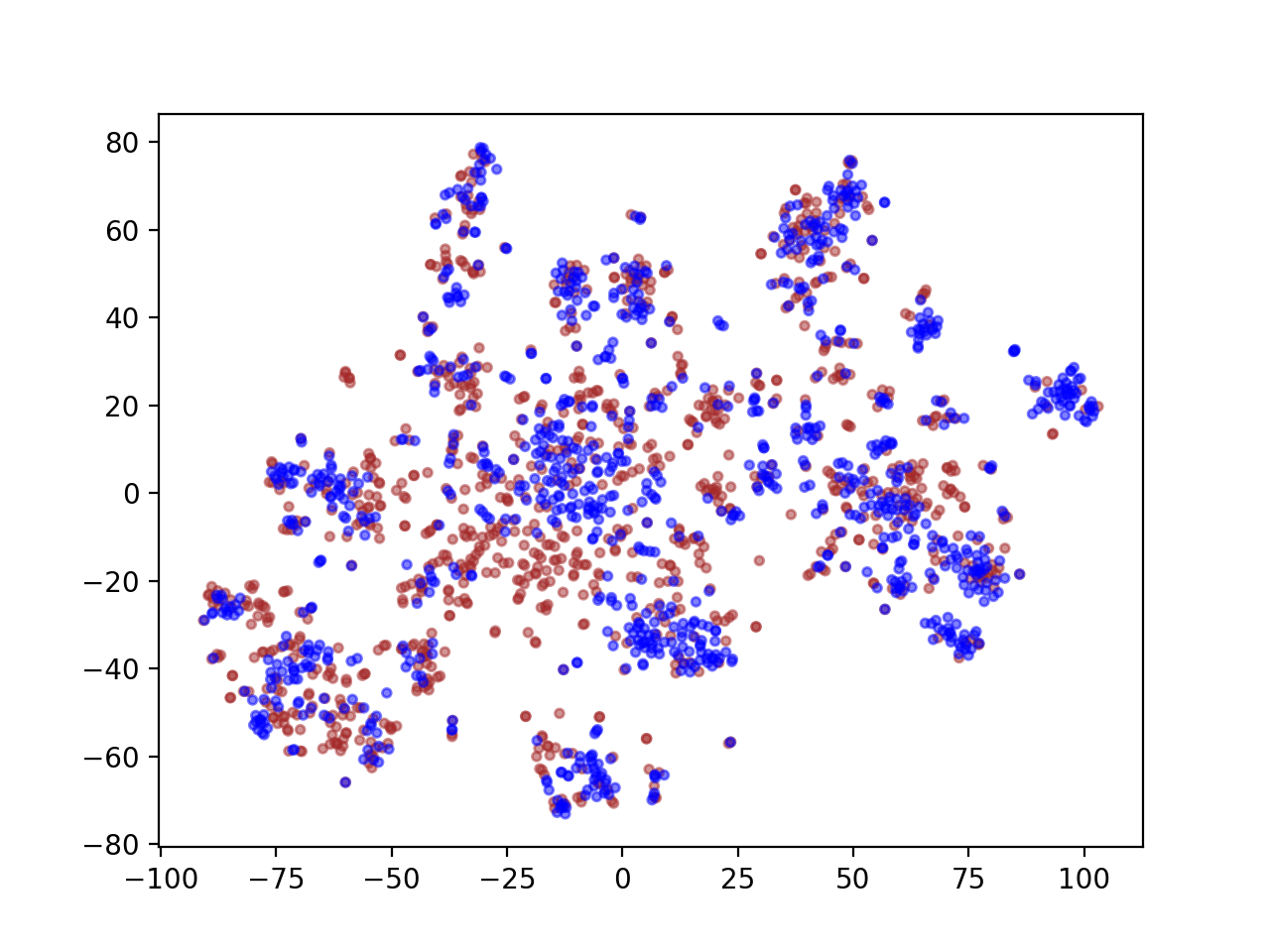}
    \caption{Embeddings of functionality-specific utterances of different origin. Brown points are from human-annotated data, and blue points are from SSL augmented data.} 
    \label{fig:musicalarm_tsne}
\end{figure}

Figure \ref{fig:musicalarm_tsne} depicts utterance embeddings for an Alexa functionality\footnote{For the figure, we randomly sampled 1K utterances from both annotated and augmented data. 
Sentence embeddings are obtained from $88.5$ million utterances in production as described in \cite{pagliardini2017unsupervised}. For visualization we used t-SNE \cite{maaten2008visualizing}.}. 
The figure shows that 
augmented data is scattered and fills in the gaps between annotated data. 
What we aim in this work is to prioritize augmentation samples given already existing training data, so that we can fill in the distribution more efficiently. 

\label{diversity-ssl}

\section{Paraphrase Detection}

\subsection{Paraphrase Embedding} 
\label{ssec:embedding}
	To select our paraphrase embedding models, we experimented with the set of models proposed in \cite{wieting2015towards}. In that work, six models were proposed, varying in terms of expressiveness and complexity. 
	The first model only learns a word embedding matrix $W_w$, and sequences are embedded by averaging over the learned word embeddings:
	\begin{equation}
	g_{para-phrase}(x) = \frac{1}{n} \sum_{i}^{n} W_w^{x_i}
	\end{equation}
	The second model adds a projection layer to the first model and learns a second weight matrix 
	in addition. 
	\begin{equation}
	g_{para-proj}(x) = W_c(\frac{1}{n} \sum_{i}^{n} W_w^{x_i}) + b_i
	\end{equation}
	
	The third model tested is the Deep Averaging Network (DAN) of \cite{iyyer2015deep}, which generalizes the second model above to variable depth as well as nonlinear activation functions. The fourth model is a standard RNN. 
	\begin{align}
	h_t &= f(W_xW_w^{x_i} + W_hh_{t-1} + b) \\
	g_{RNN}(x) &= h_{-1}
	\end{align}
	The fifth proposed model is an identity-RNN proposed by the authors \cite{wieting2015towards}.
	In an identity-RNN, the weights are all initialized to the identity matrix, and the activation function is the identity function. 
	Before initialization, the identity-RNN reduces to the first model above and averages the word embeddings, but the expectation is that during learning the model learns semantic information to improve upon the word averaging baseline.
	The final model is an LSTM network \cite{hochreiter1997long}.

Learning the parameters for each model involves minimizing an 
	objective function with a margin~\cite{wieting2015towards}.
		\begin{align}
	&\min_{W_c, W_w} \frac{1}{\vert X \vert} \big( \sum_{\langle x_1, x_2 \rangle \in X} \max(0, \delta - \cos (g(x_1), g(x_2)) \\
	&+ \cos(g(x_1), g(t_1))) + \max(0, \delta - \cos (g(x_1), g(x_2)) \\ 
	&+ \cos(g(x_2), g(t_2))) + \lambda_c \vert \vert W_c \vert \vert ^2 + \lambda_w \vert \vert W_{w_0} - W_w \vert \vert ^2   \big) 
	\end{align}
	Where $\delta$ is the margin, $g()$ is the embedding function, $\gamma_c$ and $\gamma_w$ are regularization parameters, $W_{w_0}$ is the initial word embedding matrix, and $t_1$ and $t_2$ are negative examples.
	
To construct negative pairs for training ($t_1$ and $t_2$),  
	we selected the most similar non-paraphrase to an utterance, in paraphrase pairs $x_1$ and $x_2$ respectively,   
	from a training minibatch.
	For each model we learned paraphrase embeddings using the PPDB-S dataset \cite{pavlick2015ppdb}, which includes 1.5 million paraphrase pairs.
	We trained each of the above models and evaluated them on the PPDB human-ranking task \cite{pavlick2015ppdb} and found that the word-averaging method performed best in our setup. 
	Therefore, we selected the word averaging model as the embedding layer for our downstream paraphrase detection task. 

\subsection{Paraphrase Data}
In this work, a group of utterances that share the same functionality in our dialog system are defined as paraphrase. In order to group such utterances, we rely on their NLU annotation. 

	We trained the paraphrase detection model on a set of internally-collected utterance pairs.
In order to find and manually annotated paraphrase pairs, we look into 
	the difference in time between the two utterances.
	If a request fails, users may immediately attempt to rephrase the original request so that the model can correctly process the request the second time around. As this fail-success pair may also include other errors (e.g. ASR error), our human annotators annotated whether it is indeed a paraphrase pair. 
	This data consisted of 9.1K utterance pairs, with roughly one third of the pairs annotated as paraphrases.

	Because the human annotated data is 
	relatively small, we build another corpus using utterances in NLU training data, as shown in \cite{cho-etal-2019-paraphrase}. 
	We collected 
	utterance pairs
if they had the same domain, intent, and number and types of slots in their NLU annotation.
	We then removed the slot entities so that we could compare the slot types, not specific slot entities.
	For example, we masked a paraphrase pair, 
	as in \textit{play Artist - I want to listen to Artist}. 
	After the pairs were collected, we added back in slot entities so that they would match between the two utterances by randomly selecting entities from an internal catalog for each slot type. 
	After this step, each utterance pair was considered a positive example for the paraphrase detection task.
	To include generated negative examples into training, we randomly selected two utterances from our entire set of utterances. 
In addition to this, we added negative examples where carrier-phrases are the same but different entities are randomly sampled (e.g. \textit{play Adele} - \textit{play Ed Sheeran}). 
	As a result, we added an additional $2.7$ million examples to our training set.

\subsection{Paraphrase Detection Model}
We built a paraphrase model to output a score indicative of the likelihood that a pair of utterances are paraphrases.
	The model is defined as follows:
	\begin{align}
	&e_i = g(x_i) \\
	&h = [ e_1, e_2, \vert e_1 - e_2\vert, e_1 \times e_2] \\
	&p(\text{para}(x_1, x_2)) = \sigma(h) 
	\end{align}

	In this model, we embed both utterances using the selected paraphrase embedding model (Section \ref{ssec:embedding}).
	We then combine the embeddings to form an input representation 
	by concatenating the embedding for each utterance, the element-wise difference between the two utterances, and the element-wise product between the two.
	This representation is then passed through a fully-connected neural network with a single output, representing the probability that the two utterances are paraphrases.
	We experimented with a number of hidden layers and activation functions and found that using two 
	$100$-dimension hidden layers with ReLU activation \cite{nair2010rectified} gave the best performance. 

\subsection{Data Selection using Paraphrase Model}

Our method to select utterances from the augmentation data we obtained from SSL is shown in Algorithm \ref{para_greedy}. 
\begin{algorithm}[h]
\small{
\caption{Greedy algorithm for diversity}\label{para_greedy}
\begin{algorithmic}[1]
\Require{Annotated $A = (a_1, a_2, ..., a_m)$}
\Require{Augmented $U = (u_1, u_2, ..., u_l)$}
\Require{Batch size $s = \vert U \vert \times 5 / 100$}
\Require{Augmentation budget} $k = p \times \vert U \vert, 0 < p \leq 1 $
\State $S \gets \emptyset$
\While{$\vert S \vert < k$}
\State $\mathcal{D} \gets \emptyset$ 
    \For{$u_i \in U$} 
        \State{$pr(u_i) \gets \max_j \text{para}(u_i, a_j)$ for $a_j\in A$}
    \EndFor
    \While{$\vert \mathcal{D} \vert < s $}
        \State $i^* = {\rm argmin}_{i} pr(u_i)$
        \State $\mathcal{D} \gets \mathcal{D} \cup \{u_{i^*}\}$
        \State $U \gets U - \{u_{i^*}\}$
    \EndWhile 
    \State $A \gets A \cup \mathcal{D}$
    \State $S \gets S \cup \mathcal{D}$
\EndWhile
\State use $S$ for training
\end{algorithmic}
}
\end{algorithm}

In order to select utterances from the augmented data $U$, we measured similarity between $U$ and the annotated data set $A$. 
In steps $4$ and $5$, 
we find the best matching paraphrase 
for each utterance in $U$ to utterances in $A$. 
In each iteration of selection, we choose $5$\% of the utterances in $U$, that are most dissimilar to $A$ to prioritize diverse utterances for training. We split the augmentation budget into smaller batches to successively expand the support of the augmented data. For instance, after the first iteration, we augment data $\mathcal{D}$, from $U$ that is least similar to $A$, into $A$. In subsequent iterations, we search for data to augment that is both dissimilar to the original $A$ as well as the augmented $\mathcal{D}$. 

In our preliminary experiment, we compared multiple sampling schemes. 
First, we considered prioritizing utterances for selection by using either the ${\rm argmax}$ or ${\rm argmin}$ operator in line $7$ of the Algorithm \ref{para_greedy}. 
As expected, increasing diversity with ${\rm argmin}$ was most effective.
Second, we evaluated how to handle duplicate utterances which can occur since both sets $A$ and $U$ are sampled from production traffic. In one approach we refrained from selecting duplicate utterances as long as there are other candidates in the batch. In another approach, we allowed duplicate occurrences while sampling. Results showed that the best performance was achieved when we sampled unique utterance first, prioritizing diversity. 
\label{paraphrase-detection} 

\section{Benchmarks Setup}\label{benchmarks} 
%
%

To simulate the effect of SSL as well as the efficiency improvement afforded by data selection using diversity, we established benchmarks. 
For each benchmark, 
we created 
different versions of the NLU model with varying amount of live training data on the target functionality in order to simulate the development cycle of it. 
Functionalities with varying complexity are chosen in order to better represent the diversity within 
our NLU model. 

\begin{table}[b]
\small{
\begin{center}
\begin{tabular}{c|c|c|c|r|r}
\hline 
Funct. & Domain & \#Int. & \#Slot & Annt. & Test\\ \hline 
Announce & Comms. &1 & 17 (1) & 4.4K & 1.3K \\ 
Quotes & Info & 1& 12 (5) & 4.1K & 1.4K \\ 
Playlist & Music & 2& 32 (0) & 5.2K & 1.9K \\ 
Alarms & Music & 2& 51 (4)  & 10.6K & 2.7K\\ 
Chat & General & 1& 1 (1) & 1.3K & 2.7K \\ 
\hline 
\end{tabular}
\end{center}
}
\caption{\label{Benchmarks_table} Data statistics for NLU functionalities.}
\end{table}

The baseline for each benchmark only includes synthetically created training data from FST for the target functionality. On top of that, 10\%, 20\%, 50\%, 80\% and 100\% of the annotated live training data for target functionality is added to simulate the development cycle. 
In this paper, we will refer to this as \textit{annotation increment}. On each annotation increment, SSL is carried out and a training data augmentation set is selected, as described in Section \ref{ssl-nlu}. Finally, paraphrase detection is employed to sub-select the SSL augmentation data. 

\begin{table*}[ht]
\begin{center}
\small{
\begin{tabular}{c|l|c|r|c|c|c}
\hline 
Increment & System    & Announce & Quotes    & Playlist   & Alarms   & Chat \\ \hline 
0\%                     & Baseline & 41.05     & 107.44    & 51.33 & 22.75     & 30.86 \\ \hline 
\multirow{2}{*}{10\%}   & Annotation    & 34.05     & 77.02     & 40.61 &  16.01    &  34.03 \\ \cline{2-7} 
                        & + SSL     & \textbf{27.16}     & \textbf{67.45}     & \textbf{33.32} & \textbf{15.12}     & \textbf{26.73} \\ \hline 
\multirow{2}{*}{20\%}   & Annotation    & 31.78     & 70.02     & 37.89 & 14.28     &  34.72 \\ \cline{2-7} 
                        & + SSL     & \textbf{24.75}     & \textbf{59.98}     & \textbf{32.38} & \textbf{13.37}     & \textbf{28.78}\\ \hline
\multirow{2}{*}{50\%}   & Annotation    & 29.11     & 57.95     & 31.11 & 11.65     & 31.70\\ \cline{2-7} 
                        & + SSL     & \textbf{23.71}     & \textbf{50.43}     & \textbf{25.78} & \textbf{10.97}     & \textbf{24.98} \\ \hline
\multirow{2}{*}{80\%}   & Annotation    & 28.08     & 50.70     & 26.92 & 11.20     & 31.19\\ \cline{2-7} 
                        & + SSL     & \textbf{24.85}     & \textbf{43.40}     & \textbf{23.38} & \textbf{10.62}     & \textbf{24.54}\\ \hline
\multirow{2}{*}{100\%}  & Annotation    & 28.36     & 50.82     & 26.13 & 10.90     & 30.30 \\ \cline{2-7} 
                        & + SSL     & \textbf{23.40}     & \textbf{41.40}     & \textbf{21.90} & \textbf{10.40}     & \textbf{22.73}\\ \hline
\end{tabular}}
\end{center}
\caption{\label{Benchmarks_data_stat} The impact of SSL on NLU Performance for five functionalities. Numbers are reported in SER. }
\end{table*}

\begin{table}[ht]
\caption{\label{res-all} SER score of five functionalities for each annotation increment using different selection schemes. The best selection mechanism is shown in bold letters.} 
\small
\begin{center}
\begin{tabular}{c|c|c|c|c|c|c}
\hline 
Incr. & Select & Annc. & Quotes & Playlist & Alarms & Chat \\ \hline 
\multirow{5}{*}{10\%}
& \texttt{ALL} & 27.16  & 67.45             & 33.32             & 15.12             & 26.73 \\ 
& Para  & \textbf{28.14} & \textbf{67.10}   & \textbf{33.82}    & \textbf{14.91}    & \textbf{27.59} \\
& Rand  & 29.25          & 68.07            & 34.56             & 15.45             & 28.72 \\
& Sub-m.& 28.95          & 68.04            & 34.44             & 15.05             & 28.35\\
& Uniq  & 28.96          & 67.42            & 34.29             & 15.00             & 27.90 \\\hline 
\multirow{5}{*}{20\%}
& \texttt{ALL} & 25.75  &  59.98            & 32.38             & 13.37             & 28.78  \\
& Para  & \textbf{26.59} & 61.37            & 32.34             & 13.51             & \textbf{28.84} \\
& Rand  & 27.84          & 61.35            & 33.37             & 13.55             & 29.87 \\
& Sub-m.& 27.05          & \textbf{60.75}   & 32.11             & \textbf{13.33}    & 29.62\\
& Uniq  & 26.92          & 60.97            & \textbf{31.70}    & 13.57             & 29.22 \\\hline 
\multirow{5}{*}{50\%}
& \texttt{ALL} & 23.71  & 50.43             & 25.78             & 10.97             & 24.98 \\
& Para  & \textbf{24.40} & 50.41            & \textbf{25.38}    & 11.11             & \textbf{26.16} \\
& Rand  & 24.99          & 51.62            & 26.18             & 11.17             & 27.57\\
& Sub-m.& 24.69          & 50.43            & 25.76             & \textbf{11.04}    & 26.56 \\
& Uniq  & 24.64          & \textbf{50.23}   & 25.63             & 11.19             & 27.06 \\\hline 
\multirow{5}{*}{80\%}
& \texttt{ALL} & 28.08  & 43.40             & 23.38             & 10.62             & 24.54 \\
& Para  & \textbf{24.52} & \textbf{44.00}   & 23.41             & \textbf{10.61}    & \textbf{25.31} \\
& Rand  & 24.86          & 44.87            & 24.13             & 10.74             & 26.58\\
& Sub-m.& 24.85          & 44.25            & \textbf{23.38}    & 10.61             & 25.72 \\
& Uniq  & 25.35          & 44.03            & 23.42             & 10.63             & 25.79 \\\hline 
\multirow{5}{*}{100\%}
& \texttt{ALL} & 23.40  & 41.40             & 21.90             & 10.40             & 22.73 \\
& Para  & \textbf{23.41} & 42.28            & 21.98             & 10.41             & \textbf{22.80} \\
& Rand  & 24.12          & 43.05            & 22.47             & \textbf{10.35}    & 24.20 \\
& Sub-m.& 23.86          & 42.61            & 22.17             & 10.43             & 23.57 \\
& Uniq  & 23.51          & \textbf{42.06}   & \textbf{21.65}    & 10.48             & 23.01 \\\hline 
\end{tabular}
\end{center}
\end{table}

\section{Experimental Setup}

The NLU model we used for this experiment has three statistical models; domain classifier (DC), intent classifier (IC), and named entity classifier (NER).

For this experiment, we used maximum entropy (ME) classifiers for DC and IC and conditional random fields for NER modeling. We used $n$-grams extracted from training data as features for the models. 
A detailed description of our NLU system can be found in \cite{su2018re}.

NLU performance is measured in 
Slot Error Rate (SER), as described in ~\cite{makhoul1999performance}. SER is obtained by comparing NLU hypothesis and reference and counting slot errors in substitution (S), insertion (I), and deletion (D). Intent error is treated as a slot substitution. We obtain SER as follows: 
\begin{equation}
    SER = \frac{S+I+D}{S+D+C}
\end{equation}
where $C$ denotes the number of correct slots/intent.

\subsection{Functionalities and Augmentation for SSL}

As described in Section \ref{benchmarks}, we experiment with augmentation data selection on five functionalities.

Table \ref{Benchmarks_table} summarizes characteristics of each functionality, the number of utterances of annotation data at 100\% increment and test data. Characteristics of each functionality are represented by their domain, number of intents and slots that the functionality covers. We also show how many new slots (in parentheses) are introduced for named entity modeling by this new functionality. Each functionality covers one to two NLU intents in one NLU domain. We can also see that depending on the functionality, slot modeling complexity is drastically different. 
For example, it can be more challenging for models to learn a new functionality that does not bring any new slots than learning one with new additional slots, as the model needs to re-define and learn existing slots for a new functionality.

For each annotation increment, we halve the corresponding augmented data using our diversity techniques. 
Our goal is to halve the amount of augmentation data while achieving comparable NLU performance, compared to using all augmentation data. 

Our model performance is not deterministic given retraining with the same training data, because of parallel computing and subsequent randomization. Thus, in this work we report averaged performance of five times of model training.  

\subsection{Comparative Systems} 
\label{baseline_description} 
In SSL experiments, baseline uses no augmentation data for training. We will show how much NLU accuracy improvement can be obtained by applying the SSL technique. 

In experiments for data selection using diversity, the baseline uses all augmentation data. 
As comparative systems, we explored four systems. 

Experiments \textit{random} (shown as rand) demonstrate whether we can achieve comparable NLU performance by randomly selecting half of the augmentation data for an efficient training. 

We also explored data selection based on \textit{sub-modular optimization} (shown as sub-m.). We use feature-based submodular optimization \cite{kirchhoff2014submodularity} to select a subset (with a budget 50\% of the original set) from SSL augmentation data. In our experiments, we use $2$-$4$ $n$-grams as features with $1.0$ weight, tf-idf score as the modular function, and square root as the concave function. For submodular maximization, we use a lazy greedy algorithm \cite{leskovec2007cost, minoux1978accelerated}.     

Finally, we aim to provide another comparative system where the augmentation data is \textit{uniqued} (shown as uniq) and used for training. As the data size reduction of augmentation data would vary by functionality and by annotation increment, an exact comparison to 50\% data selection is not feasible by simply taking unique utterances. For comparison, we 
match the same amount of data selection, either by selecting further utterances randomly out of the unique utterances, or randomly taking a subset of the unique utterances. 
\section{Results}
\subsection{SSL Augmentation Results}

Table \ref{Benchmarks_data_stat} shows the impact of the feature-specific SSL. 
The row of Increment 0\% shows the NLU functionality performance in SER when there is no annotated data from live traffic used for training. 

We show that as the increment increases, we were able to obtain more annotation data for the functionality, improving the NLU performance gradually. By finding the funtionality-relevant utterances in given live traffic utterances and applying the SSL approach as described in Section \ref{ssl-nlu}, we could consistently obtain further improvements SER. 

Another noticeable thing is that we can achieve NLU performance of 100\% annotation increment in a much earlier annotation increment, with SSL. For example, without using SSL, Announce functionality's SER goes down to 28.36 at the annotation increment 100\%. We can easily achieve this level of performance by using SSL at the annotation increment 10\%. For functionalities such as Quotes, Playlist, and Alarms, we can achieve the 100\% annotation increments' NLU performance by using SSL at the annotation increment 50\%. This shows that via SSL we can offer significantly better NLU performance to users, with significantly smaller manual annotation efforts. 
Overall, the results show that 
we can significantly improve NLU performance of the target functionality 
using automatic augmentation across all functionalities and annotation increments.

\subsection{Data Selection for Diversity}

Table \ref{res-all} shows NLU performance for five functionalities given annotation increment and selection mechanisms. The best selection mechanism is shown in bold letters. 

Numbers in the row of \texttt{ALL} represent NLU performance on the functionality test set when all augmentation data is used for the training, without any selection. 
In each row for each annotation increment, 
we show the performance of different selection methods (paraphrase, random, sub-modular and unique) which were applied to select  
half of the 
augmentation data. The numbers are reported in SER. 

We performed a ranking based test to compare different methods to select the data. Details of the significance test can be found in 
\cite{friedman1940comparison, nemenyi1962distribution}. The test is applied for all functionalities and their all increments. When comparing four selection schemes, we learned that with 90\% confidence the paraphrase detection model is better than any of the other selection methods. 

We observe that selecting half of the augmentation data using paraphrase model does not cause a drastic performance drop compared to using all of the augmentation data---changes range from a $1.34\%$ relative decrease to a $3.61\%$ relative increase in SER. Further, a paired t-test \cite{koehn2004statistical} shows that for 7 out of 25 experiments shown in Table \ref{res-all}, training on all augmentation data is equivalent with $95\%$ confidence to training on 50\% augmentation data using paraphrase-based selection. Thus, we conclude that using greedy selection with paraphrase detection gives the best consistent performance when selecting half of the augmentation data for semi-supervised learning.

\section{Conclusions and Future Work}

In this work, we showed that using functionality-specific SSL can greatly improve performance of production NLU systems. We explored multiple approaches for {\it efficient} SSL by optimizing diversity of augmentation data. Experiments showed that we can reach comparable performance by selecting half of the augmentation data based on diversity measures compared to using all augmentation data. 

Future work includes applying diversity-measure-based utterance clustering and paraphrase-based strategies for utterance generation and augmentation. Also, we would like to investigate how the suggested SSL approach performs on the state-of-the-art approaches. 

\bibliographystyle{IEEEbib}
\bibliography{asru}
\end{document}